\documentclass[letterpaper]{article}
\usepackage[preprint]{aaai2027}
\usepackage[hyphens]{url}
\usepackage{graphicx}
\usepackage{natbib}
\usepackage{caption}
\usepackage{amsmath,amssymb,amsfonts}
\usepackage{booktabs}
\usepackage{array}
\newcommand{\branchten}{branch-within-10}
\newcommand{\topone}{top-1}
\newcommand{\toptwo}{top-2}
\newcommand{\norm}[1]{\left\lVert #1 \right\rVert}

\title{Low-Margin Branch Points in Deterministic LLM Deliberation}
\author{Hajime Shimao\textsuperscript{1}, Warut Khern-am-nuai\textsuperscript{2}}
\affiliations{
\textsuperscript{1}Penn State University\\
\textsuperscript{2}McGill University
}

\begin{document}
\maketitle

\begin{abstract}
Deterministic multi-agent LLM deliberation can exactly replay a fixed transcript, yet exact replay alone does not test local resistance to controlled internal interventions. Because each response becomes context for later agents, a changed greedy token can redirect a transcript-mediated execution. We present a white-box audit of this local failure mode in a five-agent committee under greedy decoding. We prospectively select low- and high-margin token positions, require exact no-op replay, inject norm-relative residual displacements, and measure first transcript divergence within ten generated tokens. Low-margin positions are more branch-sensitive than high-margin controls, while matched margin-reducing gradient directions branch substantially more often than margin-increasing controls. We then continue selected realized branches through the committee to distinguish textual persistence from preference and final-decision changes. The contribution is a controlled audit for local branch susceptibility in this specified protocol, not an estimate of deployment frequency, decision correctness, or a phenomenon unique to multi-agent systems.\end{abstract}

\section{Introduction}

It has been a common design for LLM deliberation to let multiple agents exchange messages in a shared conversation before making a joint recommendation or selecting an answer~\citep{wu2023autogen,chen2023agentverse,li2023camel,hong2023metagpt}. Such systems are usually motivated by the expectation that agents can compensate for one another's mistakes. That expectation does not remove dependence on earlier messages, however. Once an agent emits a token, it is stored in the transcript and becomes available to the agents that act next. Differences that appear minor at the point of generation can consequently influence later parsing and later rounds of reasoning.

Exact replay answers a limited question: does this initial state lead to the same transcript again? It does not show what happens after a small change to that state. That distinction matters because reproducibility and local stability are different properties~\citep{lorenz1963deterministic,strogatz2018nonlinear}. A deliberating LLM committee is especially awkward to analyze in this respect. Its state changes through continuous transformer computations, but its execution also contains discrete events, including argmax choices, parser updates, transcript writes, and agent transitions. The issue is therefore whether a small perturbation can trigger one such event and alter the later discussion.

The usual replay result should be read narrowly. It shows that one specified configuration can be run again without changing the transcript; it does not measure a neighborhood of internal states. This motivates a local audit question, but not a model of how often a deployment will encounter a comparable displacement. Our residual interventions are controlled stress probes for identifying branch susceptibility, rather than simulations of hardware variation, cache approximation, prompt edits, or any other natural perturbation distribution.

The practical issue is what later parts of the committee do with a changed token. A token can differ from the reference transcript without affecting anything that follows it. For example, it may be filler, or it may be present in the transcript but never enter parsed state. A different token may instead alter a premise that another agent relies on, or contribute to a changed preference and vote. We keep these cases separate when tracing a realized branch through the committee.

We study the question in a five-agent committee that discusses a three-option policy scenario for 20 rounds. The main setting keeps the labels needed to identify agents across a replay, while removing explicit instructions that assign them particular roles. Each token is generated with greedy decoding. We first reconstruct the unmodified call and compare its token ids with those of the reference transcript. An intervention is considered only after this no-op replay matches exactly, since otherwise the source of a branch is ambiguous.

At every greedy-decoding step, the model chooses the token with the largest logit. We use the gap between that token and the runner-up as a way to identify potentially sensitive positions. A large gap gives a small residual displacement little room to change the selected token. A narrow gap leaves the argmax decision closer to its boundary, so an aligned displacement may instead yield a different continuation. If that continuation is written into the shared transcript, it becomes part of the context available to the agents that act later.

Many low-margin tokens will not be important in a semantic sense. We use them because a narrow logit gap marks a place where a continuous residual change can alter the token selected by greedy decoding. That gives us a way to produce a discrete change in the transcript. The effect of that change still has to be examined in the deliberation itself. Agents may ignore the altered text, repair it, repeat it, parse it into state, or use it when forming a later preference.

To examine local stability, we first establish that the committee can be replayed exactly without an intervention. We then select token positions from the replayed trajectory and compare interventions at low-margin positions with interventions at high-margin controls. Margin-gradient interventions test whether moving toward the local decision boundary changes the token choice. Finally, we follow a selected set of realized branches after the changed text enters shared memory. The analysis concerns this audited setting: it does not estimate deployment frequency, establish that local branching is unique to committees rather than single-agent generation, or imply that every branch changes a final decision.

\section{Related Work}

\paragraph{Multi-agent LLM systems.}
Recently multi-agent LLM research has developed systems for debate, planning, review, coding, and coordinated task solving. AutoGen, AgentVerse, CAMEL, and MetaGPT treat conversation among agents as the central part of system design~\citep{wu2023autogen,chen2023agentverse,li2023camel,hong2023metagpt}. Generative-agent work provides a related example in which agents carry memory forward as they continue interacting~\citep{park2023generative}. Shen et al.~\citep{shen2025topology} causally study beneficial and erroneous information propagation under alternative communication topologies. Our experiment instead holds one committee topology fixed and audits whether a residual intervention first becomes a discrete transcript branch. This design also creates the types of dependencies studied in research about group judgment, where composition, framing, and interaction can affect a collective result~\citep{kerr2004group,hong2004groups,sunstein2002law,page2007difference,tversky1981framing}.

\paragraph{Prompt sensitivity and reproducibility.}
Studies have shown that prompt wording, formatting, and evaluation context can affect LLM behavior~\citep{sclar2023quantifying,zheng2023judging}. Works in reproducibility have therefore stressed preserving artifacts and deterministic execution~\citep{pineau2021improving,gundersen2018state}. In our setting, exact replay verifies the canonical transcript, after which controlled residual interventions ask whether selected states branch at a greedy token choice. This is a causal audit of the replayed computation, not a naturalistic perturbation-frequency study.

\paragraph{Mechanistic interventions and hybrid dynamics.}
Activation patching, causal tracing, model editing, and intervention studies relate internal model states to observed behavior~\citep{elhage2021mathematical,olah2020zoom,meng2022locating,geiger2021causal,conmy2023automated,vonrecum2026reasoning}. These works motivate the intervention primitive; our distinct unit of analysis is a replayed transcript branch and its selected continuation through shared committee memory. Classical studies of nonlinear dynamics distinguish deterministic recurrence from behavior near a trajectory~\citep{lorenz1963deterministic,eckmann1985ergodic,ott2002chaos}. Here, the relevant finite-time event is a greedy argmax change, not an asymptotic-chaos claim.

\paragraph{Auditing deliberative AI.}
The controls in our audit are designed to exclude alternative explanations for a changed transcript. The exact-replay control confirms that the reference trajectory can be reconstructed before any intervention is applied. The zero-perturbation control confirms that the intervention machinery alone does not produce a branch, and the high-margin positions serve as a comparison for the selected low-margin positions. In addition, we vary the direction of the residual intervention, so that the local argmax account can be tested. We consider this important for a committee setting, because the messages are stored and then reused in the later turns; a comparison of only the final output would not tell us where the difference first entered the run.

\section{Experimental Setup}

\subsection{Deliberation Protocol}

The main experiments use a health-financing policy task. A national commission must choose among three universal-coverage designs under a fiscal cap: single-payer with global budgets and private supplemental coverage; a public option with private plans plus subsidies; or regulated multi-payer coverage with a mandatory basic package and strict price controls. A five-agent committee deliberates for 20 rounds. Each turn produces free-form text and a structured state line with a preference vector over options A, B, and C, confidence, and tags. The transcript and structured state are carried forward, and a final decision is produced after deliberation.

After each agent speaks, its message and structured state are appended to the transcript used by the next agent. The structured state records preferences over A, B, and C, confidence, and tags. Twenty rounds of this exchange produce both a sequence of preference vectors and a final committee decision. These records are used in the continuation analysis: we check whether a transcript branch is later reflected in preferences or only changes text that the committee does not use.

The main runs use Qwen/Qwen2.5-7B-Instruct under greedy decoding~\citep{yang2024qwen25}, implemented with Hugging Face Transformers and PyTorch~\citep{wolf2020transformers,paszke2019pytorch}. Agent identifiers stay fixed across replays, but the prompts do not assign explicit roles. Additional checks use \emph{Agent 1} through \emph{Agent 5} as homogeneous labels, five further policy scenarios, and Qwen/Qwen2.5-3B-Instruct.

\subsection{Replay and Target Selection}

Each audit row begins with a reference call and a no-op reconstruction. We retain prompts, token ids, decoded text, top-k logits, \topone{}/\toptwo{} margins, parser status, structured preference vectors, and transcript hashes for both. A row is eligible for perturbation only when the reconstructed call reproduces the canonical token ids. This occurred for all 1,280 zero-perturbation rows in the main grid and for all 16 full continuations in the continuation study. The no-op rows also matched in the content-token, scenario-replication, model-size, and neutral-label checks.

We next compare canonical prompts with prompt-surface variants at the same protocol location. A hidden summary is the final-position residual vector at a recorded layer; hidden divergence is the mean pairwise Euclidean distance among the matched summaries. These raw distances are used only to rank candidate windows and select an intervention layer, not as layer-normalized evidence of a universal geometry. Mean hidden distance rises from 0.17 at layer 0 to 6.70 at layer 27, and the largest layer-27 candidate window has mean pairwise distance 12.80.

\begin{figure}[t]
\centering
\includegraphics[width=0.94\columnwidth]{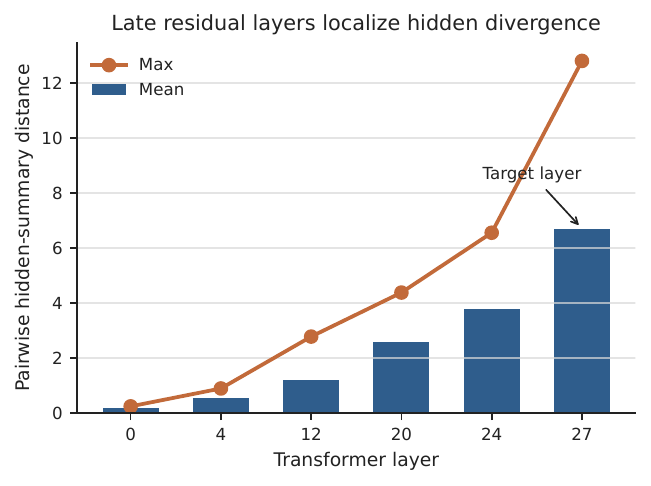}
\caption{Layer-wise raw hidden-summary divergence for canonical and prompt-surface variants at matched protocol locations. We use this localization statistic only to rank candidate windows and choose layer 27 for the subsequent causal audit.}
\label{fig:hidden}
\end{figure}

Token positions for the causal grid come from canonical logs, not from the prompt-surface comparison. In each of five main target calls, we select two low-margin tokens, one median-margin token, and one high-margin token among the first ten generated tokens; lower margin and then lower token index resolve ties. We evaluate \branchten{}: the first difference from canonical generation in the ten generated tokens strictly after the injected step. Thus \branchten{} can capture either an immediate next-token change or a delayed divergence in the local continuation; it is not an assertion that every intervention flips the token at the injection position.

\subsection{Residual Perturbations and Boundary Tests}

We inject a displacement once at the final sequence position of the post-block output of transformer layer 27 during a selected target generation step. A forward hook replaces that output $h$ with $h'=h+\Delta h$ before logits are computed; it does not directly edit an existing key--value cache. Generation then proceeds with ordinary greedy decoding and caching. A row specifies the call, token index, direction, sign, and scale. For a unit direction $v$,
\[
  \Delta h = \epsilon \norm{h}_2 v,
\]
so $\epsilon$ is relative to the pre-injection residual norm. The fixed main grid is $\{0,0.001,0.003,0.01,0.03,0.1,0.3,1.0\}$; larger values are stress probes for branch susceptibility rather than estimates of deployment noise.

The main grid includes signed directions taken from prompt-variant residual differences, seeded random directions, and orthogonal controls at the primary target calls. We also construct a direction from the local token margin. At a selected position, we teacher-force the canonical prefix, retain the layer-27 residual state $h$, and compute
\[
  m(h)=\ell_{y_1}(h)-\ell_{y_2}(h),
\]
where $y_1$ and $y_2$ are the canonical \topone{} and \toptwo{} token ids. Backpropagation gives $g=\nabla_h m(h)$. We use $-g/\norm{g}$ to reduce the canonical margin locally and $+g/\norm{g}$ to increase it.

The two gradient directions provide a matched check at the same token position. Movement that lowers the margin should make a branch more likely than movement that raises it. We compare the resulting branch rates while holding the scale fixed.

\section{Results}

\subsection{Low-Margin Tokens Branch}

The main grid completed all 10,240 planned target-call rows. Its zero-perturbation replays matched the reference tokens in all 1,280 rows. Low-margin positions produced \branchten{} in 604 of 5,120 rows (11.8\%; Wilson 95\% interval: 10.9--12.7\%); high-margin controls produced no events in 2,560 rows (0.0--0.1\%). There are twice as many low-margin rows because each target call contributes two low-margin positions and one high-margin position. The low-margin rate is higher in each of the five target calls. These intervals summarize completed, clustered intervention-dose rows; they are not population-level uncertainty over independently sampled prompts, models, or protocols.

\begin{figure}[t]
\centering
\includegraphics[width=0.94\columnwidth]{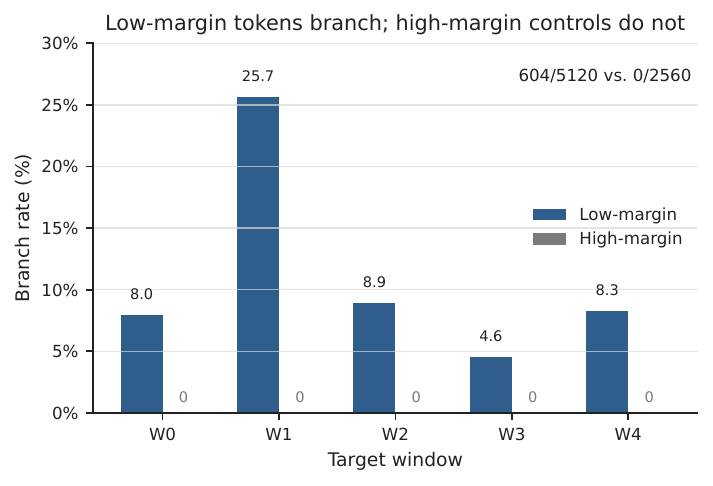}
\caption{Low-margin branch susceptibility. Low-margin positions branch within ten tokens in 604/5,120 rows; high-margin positions branch in 0/2,560 rows.}
\label{fig:low-high}
\end{figure}

The difference remains after removing the largest intervention scale. Without $\epsilon=1.0$, the low-margin rows contain 270 branches in 4,480 trials, and the high-margin rows contain none in 2,240 trials. Removing both $\epsilon=1.0$ and $\epsilon=0.3$ leaves 142 branches in 3,840 low-margin trials and none in 1,920 high-margin trials. The first switched token is usually a nearby candidate: across main-grid branch rows, its mean share in the canonical top five alternatives is 0.695.

\begin{figure}[t]
\centering
\includegraphics[width=0.94\columnwidth]{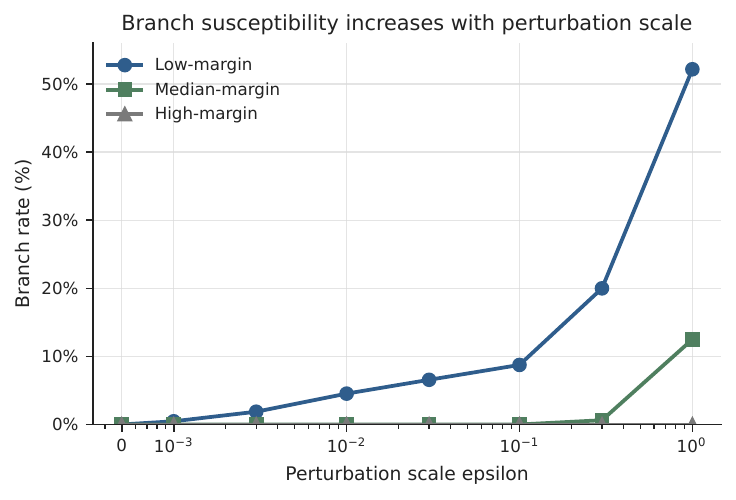}
\caption{Dose response. Branch probability rises with perturbation scale at low-margin positions, while high-margin controls remain stable over the grid.}
\label{fig:dose}
\end{figure}

Figure~\ref{fig:dose} shows the same pattern across the tested scales. Low-margin branch rates increase as the residual displacement grows. The high-margin positions remain unchanged over that range. The result is therefore concentrated at token choices with a narrow canonical margin, rather than appearing whenever a late-layer residual state is perturbed.

\subsection{The Effect Follows Boundary Geometry}

For the boundary test, we compare the two signs of the local margin gradient at the same selected token position and scale. The margin-reducing direction crosses the \branchten{} threshold in 0.433 of the 360 prospective rows; the margin-increasing direction does so in 0.025. The expected first-order effect of these signs motivates this diagnostic; its empirical role is to test whether the directional contrast survives the full replayed transformer computation and greedy continuation, rather than to claim that $-g$ is an unexpected direction. Figure~\ref{fig:gradient} shows the result.

The retrospective results are consistent with this comparison. Among the main-grid rows, median alignment with the margin-reducing gradient is 0.0138 for branch rows and approximately zero for non-branch rows. The Spearman correlation with branch status is 0.173 for alignment and 0.311 for the combined score that includes displacement size and canonical margin.

\begin{figure}[t]
\centering
\includegraphics[width=0.96\columnwidth]{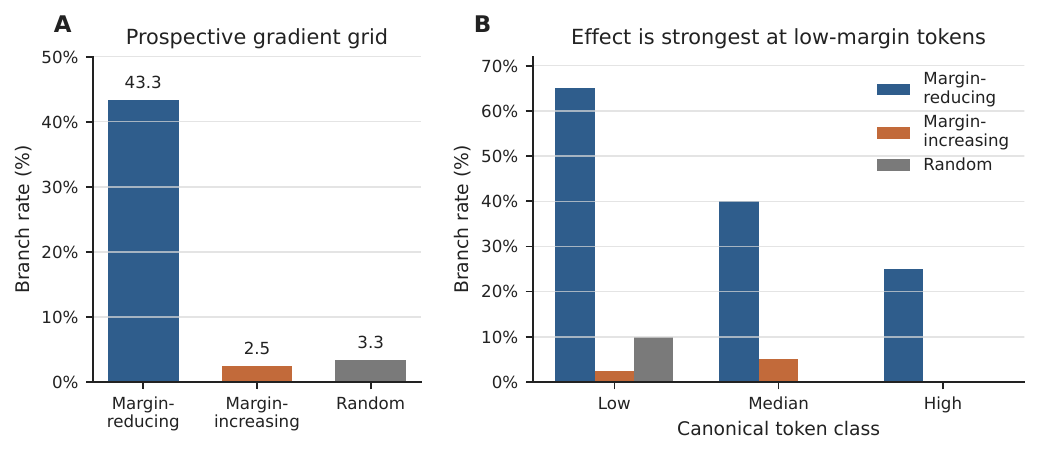}
\caption{Branch rates for the two signs of the local margin gradient. Reducing the canonical \topone{}/\toptwo{} margin produces substantially more branches.}
\label{fig:gradient}
\end{figure}

In many of these runs, the first changed token is a canonical top alternative. The intervention changes a close next-token comparison, after which the remainder of the response is generated from a different prefix.

\subsection{Realized Branches Can Propagate}

For the continuation runs, we reconstruct the committee state immediately before a selected target call. We replay that call with the selected residual displacement, parse the resulting response, add the response and structured state to the transcript, and continue deterministic generation through round 20 and the final decision. When available, the same continuation block includes a no-op replay, a high-margin control, a random non-branch control, and a random branch control.

The 16 no-op continuations reproduce all downstream calls and the final state exactly. The high-margin and random non-branch controls also have median preference-distance AUC equal to 0. The 15 low-margin branch continuations all produce a continuing textual difference; four are followed by a preference or decision change, and three end with a different final decision. Their median preference-distance AUC is 0.121, with a maximum of 0.810. This is a selected, branch-conditioned continuation set: it demonstrates that uptake by later agents and the parser can occur, not its prevalence among all target positions or branches.

\begin{figure}[t]
\centering
\includegraphics[width=0.96\columnwidth]{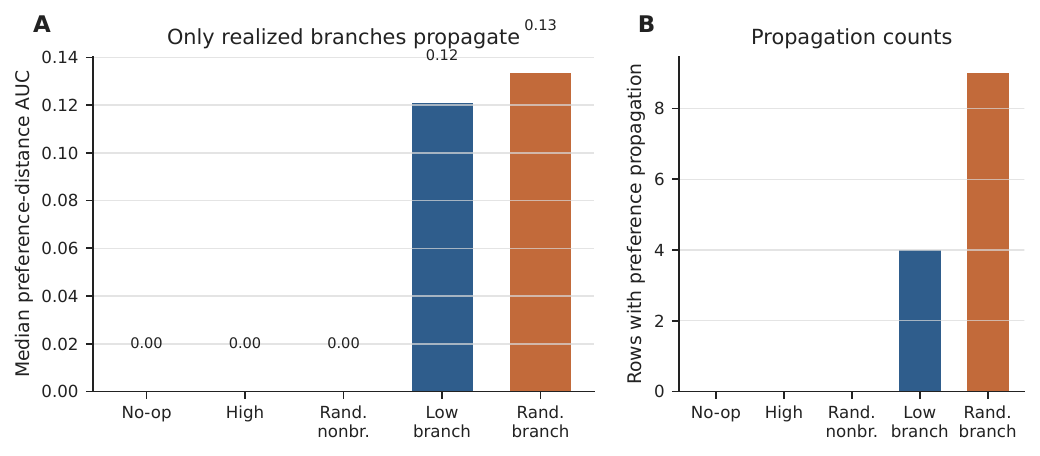}
\caption{Continuation results. No-op, high-margin, and random non-branch runs have zero median downstream preference-distance AUC. Realized low-margin and random branch continuations can produce later differences.}
\label{fig:propagation}
\end{figure}

The branch outcomes are not all the same. Some remain a change in the generated text, while others enter the parsed preference state or change the decision. Random-direction branch controls can also propagate when they produce comparable text changes. The local margin affects whether a branch is realized; what happens after that branch depends on how later agents and the parser use the altered transcript.

\subsection{Robustness Checks}

A content-token grid addresses the concern that high-margin controls in the first-ten-token grid may overrepresent punctuation or structural tokens. We select alphanumeric content-bearing tokens from the first 30 generated tokens and repeat the low/high comparison. The grid completes 5,040 rows, with 720 zero-perturbation rows and zero replay mismatches. Low-margin content tokens branch in 136/1,260 rows (10.8\%), while high-margin content tokens branch in 0/1,260 rows. Among content-token \branchten{} rows, 62.8\% switch to the canonical top-2 alternative and 87.6\% to a canonical top-5 alternative.

The additional scenario grid covers five policy scenarios with two target calls per scenario. It completes 5,460 rows; all canonical replay checks and all 780 zero-perturbation rows pass. Low-margin tokens branch in 542/3,640 rows (14.9\%), while high-margin tokens branch in 3/1,820 rows (0.16\%). Low-margin tokens are more susceptible in all 10 scenario-replication target calls. The three high-margin branches occur in one target call and only at the largest scale.

\begin{figure}[t]
\centering
\includegraphics[width=0.98\columnwidth]{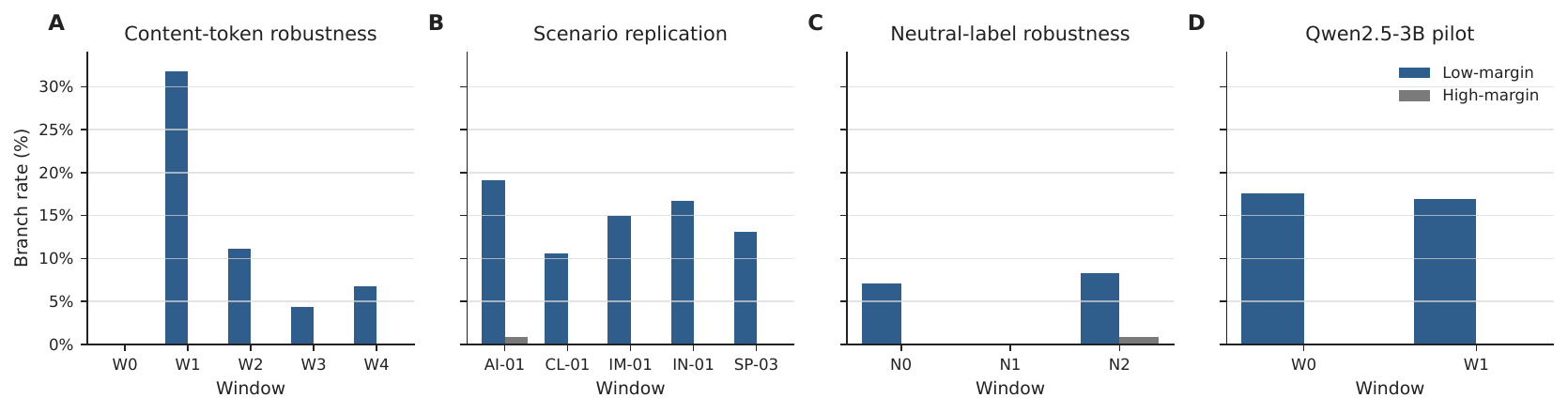}
\caption{Robustness checks. Low-margin tokens remain more branch-sensitive for content-bearing tokens, additional scenarios, a smaller checkpoint, and neutral labels.}
\label{fig:robustness}
\end{figure}

A compact Qwen2.5-3B-Instruct pilot completes 462 rows with zero replay mismatches. Low-margin tokens branch in 53/308 rows (17.2\%), while high-margin tokens branch in 0/154 rows. A neutral-label ablation completes 2,730 rows, with zero replay mismatches among 390 zero-perturbation rows. Low-margin neutral-label tokens branch in 56/1,092 rows (5.13\%), while high-margin tokens branch in 3/1,092 rows (0.27\%). These checks are target-call replications rather than full repetitions of the boundary-gradient and continuation studies, but they support that the local effect is not only a punctuation, scenario, model-size, or role-label artifact (Fig.~\ref{fig:robustness}).

The robustness pattern should be read with its scope in mind. The additional scenario and 3B experiments broaden the local low-margin result, while the content-token grid addresses a specific alternative explanation about punctuation-heavy high-margin controls. The neutral-label check is smaller, but it matters because agent labels can act as persistent semantic anchors in a committee transcript. Preserving the low-versus-high contrast after replacing named labels with homogeneous labels supports that the effect is not simply an artifact of role semantics.

\section{Discussion and Limitations}
This audit separates three questions that exact replay alone conflates. First, no-op replay asks whether the recorded trajectory can be reconstructed. Second, the perturbation grid asks whether a specified local intervention realizes a different greedy transcript. Third, selected continuations ask whether a realized text difference is later ignored, parsed into preference state, or reflected in a final decision. The evidence supports the first two questions strongly within this protocol and establishes the possibility, rather than a frequency estimate, for the third.

The boundary-gradient test is a mechanism check, not a claim that calculus makes the result surprising. Under a local linear approximation, a margin-reducing direction should be the most direct way to reduce the canonical top-1/top-2 gap. Its purpose here is to test that this expected local relation is realized in an end-to-end replayed transformer generation with a discrete greedy continuation, alongside exact-replay and random-direction controls. Raw margin remains a useful prospective warning signal, but it is not itself a calibrated residual-space distance to an argmax boundary.

The committee setting supplies a concrete route from a local token change to later collective state: the altered response is parsed and committed to shared transcript memory for subsequent agents. The results do not establish that local branch susceptibility is exclusive to committees, nor do they isolate an amplification effect relative to matched single-agent or no-sharing continuations. Random-direction branches can propagate when they create comparable text, so downstream uptake is not unique to a particular residual direction. The measured preference vectors are only one summary of a discussion, and a final decision can also reconverge after intermediate divergence.

Several limitations delimit the result. The full mechanism and continuation studies use one primary 7B model, one primary policy scenario, one late residual injection layer, and greedy decoding. The content-token, scenario, model-size, and neutral-label checks broaden the local low-margin contrast but do not repeat the complete mechanism-and-propagation study. The selected high-divergence windows and branch-conditioned continuations do not estimate prevalence. Nor do norm-relative stress probes model realistic numerical, cache, or hardware perturbations. Finally, the primary observations are clustered within five target calls; row-level intervals are descriptive summaries of this intervention grid.

Future work should compare matched single-agent, shared-transcript, and no-sharing protocols; repeat the full audit across models, layers, tasks, and decoding rules; and measure perturbation distributions induced by quantization, cache approximation, or hardware variation. A practical audit can meanwhile retain exact replay records, inspect margin-conditioned branch susceptibility around consequential transcript writes, and continue any realized branch to distinguish textual divergence from downstream state changes.
\bibliography{refs}

@article{lorenz1963deterministic,
  author  = {Lorenz, Edward N.},
  title   = {Deterministic Nonperiodic Flow},
  journal = {Journal of the Atmospheric Sciences},
  volume  = {20},
  number  = {2},
  pages   = {130--141},
  year    = {1963},
  doi     = {10.1175/1520-0469(1963)020<0130:DNF>2.0.CO;2}
}

@book{strogatz2018nonlinear,
  author    = {Strogatz, Steven H.},
  title     = {Nonlinear Dynamics and Chaos: With Applications to Physics, Biology,
               Chemistry, and Engineering},
  edition   = {2nd},
  publisher = {Westview Press},
  year      = {2018}
}

@article{sunstein2002law,
  author  = {Sunstein, Cass R.},
  title   = {The Law of Group Polarization},
  journal = {The Journal of Political Philosophy},
  volume  = {10},
  pages   = {175--195},
  year    = {2002},
  doi     = {10.1111/1467-9760.00148}
}

@book{page2007difference,
  author    = {Page, Scott E.},
  title     = {The Difference: How the Power of Diversity Creates Better Groups,
               Firms, Schools, and Societies},
  publisher = {Princeton University Press},
  year      = {2007}
}

@article{tversky1981framing,
  author  = {Tversky, Amos and Kahneman, Daniel},
  title   = {The Framing of Decisions and the Psychology of Choice},
  journal = {Science},
  volume  = {211},
  number  = {4481},
  pages   = {453--458},
  year    = {1981},
  doi     = {10.1126/science.7455683}
}

@inproceedings{sclar2023quantifying,
  author    = {Sclar, Melanie and Choi, Yejin and Tsvetkov, Yulia and Suhr, Alane},
  title     = {Quantifying Language Models' Sensitivity to Spurious Features
               in Prompt Design or: How I Learned to Start Worrying About
               Prompt Formatting},
  booktitle = {International Conference on Learning Representations (ICLR)},
  year      = {2024},
  url       = {https://arxiv.org/abs/2310.11324}
}

@inproceedings{zheng2023judging,
  author    = {Zheng, Lianmin and Chiang, Wei-Lin and Sheng, Ying and others},
  title     = {Judging {LLM-as-a-Judge} with {MT-Bench} and Chatbot Arena},
  booktitle = {Advances in Neural Information Processing Systems (NeurIPS)},
  volume    = {36},
  year      = {2023},
  url       = {https://arxiv.org/abs/2306.05685}
}

@misc{wu2023autogen,
  author    = {Wu, Qingyun and Bansal, Gagan and Zhang, Jieyu and others},
  title     = {{AutoGen}: Enabling Next-Gen {LLM} Applications via Multi-Agent
               Conversation},
  year      = {2023},
  eprint    = {2308.08155},
  archivePrefix = {arXiv},
  url       = {https://arxiv.org/abs/2308.08155}
}

@misc{chen2023agentverse,
  author    = {Chen, Weize and Su, Yusheng and Zuo, Jingwei and others},
  title     = {{AgentVerse}: Facilitating Multi-Agent Collaboration and
               Exploring Emergent Behaviors in Agents},
  year      = {2023},
  eprint    = {2308.10848},
  archivePrefix = {arXiv},
  url       = {https://arxiv.org/abs/2308.10848}
}

@inproceedings{park2023generative,
  author    = {Park, Joon Sung and O'Brien, Joseph C. and Cai, Carrie J. and Morris, Meredith Ringel and Liang, Percy and Bernstein, Michael S.},
  title     = {Generative Agents: Interactive Simulacra of Human Behavior},
  booktitle = {Proceedings of the 36th Annual ACM Symposium on User Interface Software and Technology (UIST)},
  year      = {2023},
  pages     = {1--22},
  doi       = {10.1145/3586183.3606763}
}

@misc{li2023camel,
  author    = {Li, Guohao and Hammoud, Hasan and Itani, Hani and Khizbullin, Dmitrii and Ghanem, Bernard},
  title     = {{CAMEL}: Communicative Agents for Mind Exploration of Large Language Model Society},
  year      = {2023},
  eprint    = {2303.17760},
  archivePrefix = {arXiv},
  url       = {https://arxiv.org/abs/2303.17760}
}

@misc{hong2023metagpt,
  author    = {Hong, Sirui and Zhuge, Ming and Chen, Junjie and others},
  title     = {{MetaGPT}: Meta Programming for Multi-Agent Collaborative Framework},
  year      = {2023},
  eprint    = {2308.00352},
  archivePrefix = {arXiv},
  url       = {https://arxiv.org/abs/2308.00352}
}

@article{pineau2021improving,
  author  = {Pineau, Joelle and Vincent-Lamarre, Philippe and Sinha, Koustuv and Larivi\`ere, Vincent and Beygelzimer, Alina and d'Alch\'e-Buc, Florence and Fox, Emily and Larochelle, Hugo},
  title   = {Improving Reproducibility in Machine Learning Research: A Report from the NeurIPS 2019 Reproducibility Program},
  journal = {Journal of Machine Learning Research},
  volume  = {22},
  number  = {164},
  pages   = {1--20},
  year    = {2021},
  url     = {https://jmlr.org/papers/v22/20-303.html}
}

@inproceedings{gundersen2018state,
  author    = {Gundersen, Odd Erik and Kjensmo, Sigbj\o{}rn},
  title     = {State of the Art: Reproducibility in Artificial Intelligence},
  booktitle = {Proceedings of the Thirty-Second AAAI Conference on Artificial Intelligence (AAAI)},
  year      = {2018},
  pages     = {1644--1651}
}

@misc{elhage2021mathematical,
  author = {Elhage, Nelson and Nanda, Neel and Olsson, Catherine and Henighan, Tom and Joseph, Nicholas and Mann, Ben and Askell, Amanda and Bai, Yuntao and Chen, Anna and Conerly, Tom and DasSarma, Nova and Drain, Dawn and Ganguli, Deep and Hatfield-Dodds, Zac and Hernandez, Danny and Jones, Andy and Kernion, Jackson and Lovitt, Liane and Ndousse, Kamal and Amodei, Dario and Brown, Tom and Clark, Jack and Kaplan, Jared and McCandlish, Sam and Olah, Chris},
  title = {A Mathematical Framework for Transformer Circuits},
  year = {2021},
  url = {https://transformer-circuits.pub/2021/framework/index.html}
}

@article{olah2020zoom,
  author = {Olah, Chris and Cammarata, Nick and Schubert, Ludwig and Goh, Gabriel and Petrov, Michael and Carter, Shan},
  title = {Zoom In: An Introduction to Circuits},
  journal = {Distill},
  year = {2020},
  doi = {10.23915/distill.00024.001}
}

@inproceedings{meng2022locating,
  author = {Meng, Kevin and Bau, David and Andonian, Alex and Belinkov, Yonatan},
  title = {Locating and Editing Factual Associations in {GPT}},
  booktitle = {Advances in Neural Information Processing Systems},
  volume = {35},
  year = {2022},
  pages = {17359--17372}
}

@inproceedings{geiger2021causal,
  author = {Geiger, Atticus and Lu, Hanson and Icard, Thomas and Potts, Christopher},
  title = {Causal Abstractions of Neural Networks},
  booktitle = {Advances in Neural Information Processing Systems},
  volume = {34},
  year = {2021},
  pages = {9574--9586}
}

@misc{conmy2023automated,
  author = {Conmy, Arthur and Mavor-Parker, Augustine N. and Lynch, Aengus and Heimersheim, Stefan and Garriga-Alonso, Adri\`a},
  title = {Towards Automated Circuit Discovery for Mechanistic Interpretability},
  year = {2023},
  eprint = {2304.14997},
  archivePrefix = {arXiv},
  url = {https://arxiv.org/abs/2304.14997}
}

@article{eckmann1985ergodic,
  author  = {Eckmann, Jean-Pierre and Ruelle, David},
  title   = {Ergodic Theory of Chaos and Strange Attractors},
  journal = {Reviews of Modern Physics},
  volume  = {57},
  number  = {3},
  pages   = {617--656},
  year    = {1985},
  doi     = {10.1103/RevModPhys.57.617}
}

@book{ott2002chaos,
  author    = {Ott, Edward},
  title     = {Chaos in Dynamical Systems},
  edition   = {2nd},
  publisher = {Cambridge University Press},
  year      = {2002}
}

@article{kerr2004group,
  author  = {Kerr, Norbert L. and Tindale, R. Scott},
  title   = {Group Performance and Decision Making},
  journal = {Annual Review of Psychology},
  volume  = {55},
  pages   = {623--655},
  year    = {2004},
  doi     = {10.1146/annurev.psych.55.090902.142009}
}

@article{hong2004groups,
  author  = {Hong, Lu and Page, Scott E.},
  title   = {Groups of Diverse Problem Solvers Can Outperform Groups of High-Ability Problem Solvers},
  journal = {Proceedings of the National Academy of Sciences},
  volume  = {101},
  number  = {46},
  pages   = {16385--16389},
  year    = {2004},
  doi     = {10.1073/pnas.0403723101}
}

@inproceedings{shen2025topology,
  author    = {Shen, Xu and Liu, Yixin and Dai, Yiwei and Wang, Yili and Miao, Rui and Tan, Yue and Pan, Shirui and Wang, Xin},
  title     = {Understanding the Information Propagation Effects of Communication Topologies in {LLM}-based Multi-Agent Systems},
  booktitle = {Proceedings of the 2025 Conference on Empirical Methods in Natural Language Processing},
  pages     = {12347--12361},
  year      = {2025},
  doi       = {10.18653/v1/2025.emnlp-main.623}
}

@inproceedings{vonrecum2026reasoning,
  author    = {von Recum, Alexander and Girrbach, Leander and Akata, Zeynep},
  title     = {Are Reasoning {LLM}s Robust to Interventions on their Chain-of-Thought?},
  booktitle = {International Conference on Learning Representations},
  year      = {2026},
  url       = {https://proceedings.iclr.cc/paper_files/paper/2026/hash/a03037317560b8c5f2fb4b6466d4c439-Abstract-Conference.html}
}

@misc{yang2024qwen25,
  author        = {Yang, An and Yang, Baosong and Zhang, Beichen and Hui, Binyuan and Zheng, Bo and Yu, Bowen and Li, Chengyuan and Liu, Dayiheng and Huang, Fei and Wei, Haoran and Lin, Huan and Yang, Jian and Tu, Jianhong and Zhang, Jianwei and Yang, Jianxin and Yang, Jiaxi and Zhou, Jingren and Lin, Junyang and Dang, Kai and Lu, Keming and Bao, Keqin and Yang, Kexin and Yu, Le and Li, Mei and Xue, Mingfeng and Zhang, Pei and Zhu, Qin and Men, Rui and Lin, Runji and Li, Tianhao and Tang, Tianyi and Xia, Tingyu and Ren, Xingzhang and Ren, Xuancheng and Fan, Yang and Su, Yang and Zhang, Yichang and Wan, Yu and Liu, Yuqiong and Cui, Zeyu and Zhang, Zhenru and Qiu, Zihan},
  title         = {{Qwen2.5} Technical Report},
  year          = {2024},
  eprint        = {2412.15115},
  archivePrefix = {arXiv},
  url           = {https://arxiv.org/abs/2412.15115}
}

@inproceedings{wolf2020transformers,
  author    = {Wolf, Thomas and Debut, Lysandre and Sanh, Victor and Chaumond, Julien and Delangue, Clement and Moi, Anthony and Cistac, Pierric and Rault, Tim and Louf, Remi and Funtowicz, Morgan and Davison, Joe and Shleifer, Sam and von Platen, Patrick and Ma, Clara and Jernite, Yacine and Plu, Julien and Xu, Canwen and Le Scao, Teven and Gugger, Sylvain and Drame, Mariama and Lhoest, Quentin and Rush, Alexander M.},
  title     = {Transformers: State-of-the-Art Natural Language Processing},
  booktitle = {Proceedings of the 2020 Conference on Empirical Methods in Natural Language Processing: System Demonstrations},
  pages     = {38--45},
  year      = {2020},
  doi       = {10.18653/v1/2020.emnlp-demos.6}
}

@inproceedings{paszke2019pytorch,
  author    = {Paszke, Adam and Gross, Sam and Massa, Francisco and Lerer, Adam and Bradbury, James and Chanan, Gregory and Killeen, Trevor and Lin, Zeming and Gimelshein, Natalia and Antiga, Luca and Desmaison, Alban and Kopf, Andreas and Yang, Edward and DeVito, Zachary and Raison, Martin and Tejani, Alykhan and Chilamkurthy, Sasank and Steiner, Benoit and Fang, Lu and Bai, Junjie and Chintala, Soumith},
  title     = {{PyTorch}: An Imperative Style, High-Performance Deep Learning Library},
  booktitle = {Advances in Neural Information Processing Systems},
  volume    = {32},
  year      = {2019},
  url       = {https://arxiv.org/abs/1912.01703}
}

\end{document}